\documentclass{article}

\usepackage[preprint]{neurips_2023}

\usepackage[hidelinks]{hyperref}       
\hypersetup{%
  colorlinks=true,
  linkcolor=blue,
  citecolor=blue,
}
\usepackage{url}            
\usepackage{booktabs}       
\usepackage{amsfonts}       
\usepackage{nicefrac}       
\usepackage{microtype}      
\usepackage{xcolor}         
\usepackage{csquotes}

\usepackage{csquotes}
\usepackage{amsmath}
\usepackage{amsthm}
\usepackage{graphicx} 
\usepackage{comment}
\usepackage{pifont}
\usepackage{dirtytalk}

\theoremstyle{definition}
\newtheorem{definition}{Definition}[section]

\newcommand{\carlos}[1]{}

\title{Kantian Deontology Meets AI Alignment: Towards Morally Grounded Fairness Metrics}

\author{%
  Carlos Mougan \\
  University of Southampton\\
  United Kingdom\\
  \texttt{c.mougan@soton.ac.uk} \\
   \And
  Joshua Brand \\
  Institut Polytechnique de Paris-Télécom Paris \\
   France \\
  \texttt{joshua.brand@telecom-paris.fr } \\
}

\begin{document}

\maketitle

\begin{abstract}
Deontological ethics, specifically understood through Immanuel Kant, provides a moral framework that emphasizes the importance of duties and principles rather than the consequences of action. Understanding that despite the prominence of deontology, it is currently an overlooked approach in fairness metrics, this paper explores the compatibility of a Kantian deontological framework in fairness metrics, part of the AI alignment field. We revisit Kant's critique of utilitarianism, which is the primary approach in AI fairness metrics and argue that fairness principles should align with the Kantian deontological framework. By integrating Kantian ethics into AI alignment, we not only bring in a widely accepted prominent moral theory but also strive for a more morally grounded AI landscape that better balances outcomes and procedures in pursuit of fairness and distributive justice.

\end{abstract}
\section{Introduction}
\carlos{add this paper \url{https://arxiv.org/pdf/2206.02897.pdf} and \url{https://arxiv.org/pdf/2105.01441.pdf}}
\carlos{model multiplicity \url{https://papers.ssrn.com/sol3/papers.cfm?abstract_id=4590481}}
\carlos{another paper in utility. In the intro they have aenumeration of different policies, check if there are some deontological principles
}
\carlos{Why the critique to deontology? For what? To get what?  a qué vienen estas críticas al deontologismo? A donde t ellevan? qué buscas con ellas? ¿Queda claro el por qué? Tambien se podría criticar de la misma manera al deontologismo que supongo que no se hace porque dices que lo dominante en AI es el utlitarismo. Aún así deberáis aclarar el por qué de la introducción de estas críticas. Quizás lo aclaras mas tarde pero hasta aquí no se ve su sentido
}
\carlos{Align related work with section 5}
\carlos{Rashomon effect and fairness? in the section of fairness-accuracy trade offs.}

Deontological ethics, with Immanuel Kant as its central figure, in particular through his seminal work \emph{Groundwork of the Metaphysics of Morals} \citep{kant1785groundwork}, emphasizes that it is through duties, or a universal set of rules, that the ethical value of actions is determined. Generally, this form of deontological ethics maintains that these duties flow from the recognition that all humans are intrinsically valuable---it is our rational nature that gives us reason to act in one way or another. This duty, or rule-based, moral theory thus aligns well with various declarations, such as the \emph{Universal Declaration of Human Rights}, that recognize a set of inalienable rights based on the individual, such as the right to life and the right to equal treatment under the law; with deontology, moral action involves respecting inescapable duties that often lay the fundamental foundation of the rights of others. While it is true that not all duties entail rights, all rights suppose a corresponding duty~\citep{watson2021right}.

Given this emphasis on evaluating actions through a set of rules or principles, deontology is most often constrasted to utilitarianism, a consequentialist ethical theory. Utilitarianism, traditionally understood through the likes of Jeremy Bentham or John Stuart Mill, focuses on the consequences of an action to determine its ethical value, considering the utility or benefit brought about by the action~\citep{sep-mill,mill1966utilitarianism,bentham1996collected}. In traditional utilitarian accounts, utility is identified as pleasure, well-being, or happiness~\citep{sep-utilitarianism-history}. The goal of moral action is therefore generally construed as increasing the amount of pleasure or happiness in our communities, with various proposed methods to measure these outcomes.

This is why utilitarian ethics is often favoured in quantitative sciences due to its emphasis on measurable outcomes, such as in the fields of economics ~\citep{posner1979utilitarianism} or public policy \citep{kymlicka2002contemporary,goodin1995utilitarianism}. It should be noted, however, that utilitarians face criticism on how one can quantify such concepts as happiness ~\citep{knutsson2016measuring}. By quantifying and comparing the various consequential benefits and harms of different actions or policies, the utilitarian approach nevertheless claims to achieve the most favourable outcome for the greatest number of individuals.

In the AI alignment field, the ethical challenge relies on ensuring that AI systems are first beneficial and then align with a diverse set of human values, encompassing moral, societal, justice-oriented, and other relevant principles. As London et al.~\cite{london2023beneficent} succinctly state,

\say{\textit{It should be axiomatic that producing some sort of benefit is a necessary condition for the ethical and responsible use of AI}.}

It is here that \emph{fairness metrics} play a vital role by helping to evaluate whether AI systems meet these prescribed values and treat a diverse set of individuals and groups affected by AI systems fairly---in other words, if they are treated in an impartial or equitable manner. As one of the primary aims of any moral theory or prescribed set of values is to establish the definition of the fair treatment of others, fairness metrics are an integral component of AI alignment.

In machine learning, researchers, in order to optimize fairness metrics, balance fairness-accuracy trade-offs to find an equilibrium between predictive performance metrics like accuracy, f1-score, AUC, mean squared errors, and utilitarian interpretations of distributive justice principles, such as equal opportunity and equal outcomes~\citep{DBLP:conf/nips/HardtPNS16,DBLP:conf/fat/HeidariLGK19}. Striving to optimize these trade-offs, researchers attempt pre-processing ~\citep{fabbrizzi2023studying,DBLP:journals/widm/NtoutsiFGINVRTP20}, in-processing~\citep{DBLP:conf/kdd/PedreschiRT08,DBLP:conf/fat/MenonW18,DBLP:conf/aistats/ZafarVGG17,DBLP:conf/pkdd/KamishimaAAS12}, and post-processing strategies~\citep{DBLP:journals/natmi/RodolfaLG21}.

The utilitarian approach currently takes primacy over other moral accounts, including deontology, in fairness metrics. Understanding that deontology is a prominent moral theory well-suited to widely accepted conceptions of human rights that recognize individual dignity and worth, and which brings important and established critiques against utilitarianism, we argue that it should be adopted into the fairness metrics approach to counter utilitarian dominance. In this paper we therefore analyze fairness metrics through a Kantian deontological approach and consider how such approach could be interpreted and introduced into the fairness metrics field.

The use of utilitarian principles in fairness metrics has already been critiqued for the overemphasis on aggregate benefits, often at the expense of individual rights and dignity. This paper highlights instances where utilitarian approaches in fairness metrics have been insufficient, suggesting that a deontological perspective, with its focus on duty and intrinsic value, could offer more balanced and ethical outcomes.

We present this paper with the following contributions:
\begin{itemize}

    \item \textit{Integration of Kantian Deontology with AI Fairness Metrics} The paper provides a novel perspective by suggesting how to integrate Kantian deontological ethics into the field of AI fairness metrics. This approach contrasts with the prevalent utilitarian focus, offering a fresh lens to evaluate ethical AI.

    \item  \textit{Critical Analysis of Utilitarianism in AI} This paper offers a critique of the utilitarian approach that is predominantly used in AI fairness metrics. This includes an examination of its limitations, particularly in AI applications with high social impact where individual rights and intrinsic value have been central. 

    \item \textit{Exploration of Practical AI Scenarios} Through two case studies on loan approval and student grading system, this paper discussed the practical implications of applying Kantian ethics in real-world AI applications. 
    


    
\end{itemize}
\subsection{Related Work: Interdisciplinary Perspectives on AI Ethics}
Discussions on how to integrate ethical values in AI systems is not unfamiliar and has garnered significant attention from both the AI research community and philosophers, leading to a thriving field of interdisciplinary work. We first present the related work of the AI research community towards ethical practices and then present the efforts of the philosophical community to align with ethical systems.

\subsubsection{AI Research Community's Developments on Ethics}

The AI research community has made substantial contributions regarding the implementation of moral norms and ethical principles in various contexts; studies such as Wallach and Allen's exploration of moral machines highlight the complexities of instilling AI with the capacity for moral decision-making~\citep{wallach2008moral}. There have also been extensive efforts to encode ethical principles into AI systems, such as Pereira et al.'s work on programming machine ethics which also addresses practical considerations on encoding principles~\cite {pereira2016programming}. More recent advancements focus on ensuring AI's alignment with human values for societal benefit, as exemplified by London and Heidari~\citep{london2023beneficent}, and by exploring how to encode societal norms and moral values into AI systems~\cite{serramia2023encoding,van2023ethics}.

One field, however, that has not received much attention regarding deontological norms is fairness metrics as its various approaches broadly remain utilitarian~\citep{barocas-hardt-narayanan,DBLP:journals/widm/NtoutsiFGINVRTP20}. In this paper, we criticize the use of utilitarian principles in fairness metrics as they overemphasize the advantage of aggregating benefits; we also highlight instances where utilitarian approaches in fairness metrics have been insufficient, suggesting that a deontological perspective, with its focus on duty and the intrinsic value of the individual, could offer more balanced and ethical outcomes. In other words, we show that current fairness evaluation metrics of AI systems is generally biased towards a utilitarian approach for defining morally permissible actions and values~\citep{DBLP:conf/asist/ChengF10}.

\subsubsection{Philosophical Contributions}
From a philosophical perspective, scholars like Gabriel have dissected the problem of AI alignment into two challenges: first, the technical challenge of encoding values and second, the normative challenge of determining which values to encode~\citep{gabriel2022toward, DBLP:journals/mima/Gabriel20}. Conceptualizing fairness as a moral value in terms of AI has also been explored~\citep{DBLP:conf/asist/ChengF10}. In our work, we consider Gabriel's two-step approach and discuss both the current state of the art of fairness metrics in supervised learning scenarios and their integration and analysis with the Kantian deontological perspective, a interdisciplinary connection that to the best of our knowledge has not been studied.

Given this gap in the existing contributions, both in AI research and philosophy, this paper aims to first argue that there is a dominant utilitarian approach to AI fairness and then propose a shift towards deontological ethics. The utilitarian approach, while practical, has historically faced criticism, particularly when it fails to consider the intrinsic value of individuals~\citep[e.g.,][]{smart1973utilitarianism,sep-utilitarianism-history}. While existing works, such as those proposing procedural approaches to fairness~\citep{DBLP:conf/aaai/Grgic-HlacaZGW18}, have begun to bridge the gap, their philosophical underpinnings often lack depth. This paper argues for reevaluating this mono-theoretic approach and advocates for incorporating deontological principles and metrics, particularly in contexts where individual rights and intrinsic human value are considered paramount.

\section{Overview of Kantian Deontology}\label{sec:overview}

It is important to first reiterate that the primary objective of this paper is not to explore the relationship between utilitarian and deontological moral perspectives with a novel perspective in favour of deontology. It does not provide an original outlook on the deontology-utilitarianism debate. Indeed, the following section that outlines the Kantian deontological critique\footnote{We chose the Kantian approach for the simple notion that if any philosopher is to be considered as essential to deontology it is Immanuel Kant.} of utilitarianism will certainly be familiar for those with a basic understanding in moral philosophy. The following section instead serves as a reminder of the two opposing normative ethics—one that is society-centred and the other that is patient-centred—and the main critiques Kantians raise against utilitarianism. It is useful to reiterate this longstanding philosophical debate from the deontological perceptive, which we will present in the following sections, given the largely mono-theoretic approach taken in fairness metrics. This paper therefore invites us to reconsider and critique the mono-theoretic approach in fairness metrics; that despite the field generally preferring utilitarianism, there are well-established critiques laid against utilitarianism by deontologists and that the deontological position can be beneficial for fairness metrics, particularly in the context of jurisdictions that place importance not only on measurable outcomes and social impact, but also on inalienable individual rights and the intrinsically valuable nature of the individual. This refers to the many jurisdictions who recognize that we have deep-seated duties to each and every individual that evade societal pressures to measure and prefer collective outcomes, such as the individual right to life and freedom from inhumane treatment, the right to privacy, and the right to equal treatment before the law. We later provide examples of where the utilitarian approach in fairness metrics has been largely considered as a failure and where, we argue, a deontological perspective could have improved the situation.

\subsection{The Categorical Imperative}
Much can be discussed regarding Kant’s moral theory, but one of the most fundamental aspects of his ethics, outlined in \emph{Groundwork} \citep{kant1785groundwork}, is the categorical imperative. This universal principle, which unconditionally binds all rational beings, provides us with a formal procedure to evaluate our actions---if actions adhere to the categorical imperative they are morally fit. Kant provides us with three primary interpretations of the categorical imperative, with the first two being the most widely discussed.\footnote{While there is disagreement on whether or not the formulations are synonymous, they nevertheless all lead us to the same set of duties~\citep{korsgaard1996sources}} The first is the \emph{Formula of Universal Law}:

\emph{Act only on a maxim that you can also will to become a universal law} (Groundwork, 4:421)\footnote{All further citations of this kind refer to \citep{kant1785groundwork}}

In other terms, this asks us to consider whether our actions can be performed universally, by all rational agents, without contradiction. Making false promises cannot be willed universally as if everyone made false promises, such as falsely promising to pay back a loan, no one would believe one another as they would distrust everyone and thus, making a false promise would be an impossible act (4:422). 

The second formulation, \emph{Formula of Humanity} is considered by Kant to be “closer to intuition” (4:436):

\say{\emph{Act in such a way that you treat humanity, whether in your own person or anyone else’s, never merely as a means, but also always as an end} (4:429)}

This provides us with the material of the moral law (4:436), the rational nature of individuals, or more specifically, our free will and humanity, that Kantian ethics aims to protect from any instrumental subordination to another individual’s, or group of individuals’, ends. This is to recognize that all persons are ends-in-themselves, meaning that they have an intrinsic value “exalted above all price” (4:434). This requires engagement with them as freely-choosing rational agents who have the capacity to decide what is valuable to them and to pursue their own projects and, importantly, when they are in a position of not fully realizing this capacity, to provide them with the necessary assistance (4:430). Returning to the case of the false promise, the categorical imperative forbids it because it hinges on deceiving another individual and forcing them into a situation for which they have not principally consented. We primarily focus on this formulation throughout the rest of this paper.

\subsection{Kantian Critique of Utilitarianism}

With this concise outline of Kant’s moral law, we now move on to consider the Kantian critique of utilitarianism. Kant was certainly not fond of taking external concepts, such as happiness, as the source of morality. In \emph{The Metaphysics of Morals} he states: 

\say{\textit{\ldots if eudaemonism (the principle of happiness) is set up as the basic principle instead of eleutheronomy (the principle of the freedom of internal lawgiving), the result is the euthanasia (the death) of all morals.}\cite[6:378]{gregor1996kant}}

Kant, however, did not directly address the utilitarian of his time, Jeremy Bentham, so it is helpful to look to contemporary Kantians, Onora O'Neill and Christine Korsgaard, who provide a direct critique of utilitarian ethics. The following provides a foundational, but certainly not exhaustive, list of critiques.
\subsubsection{Critique of the Impersonal Good}
The first critique is that utilitarianism believes that all moral actions ought to be aimed at obtaining the ‘good’ as defined as an impersonal absolute value, such as happiness or pleasure, that can be maximized. It follows that the value of any individual relies on the experience of happiness or pleasure. In more crude terms, individuals are a mere vessel for these valuable concepts. Peter Singer writes that the classical utilitarian approach, “regards sentient beings as valuable only insofar as they make possible the existence of intrinsically valuable experiences like pleasure.”\citep{singer2011}. Singer, himself a utilitarian, continues this notion that it is not the individuals that matter, but the defined utility, through the \emph{replaceability argument}. While he appreciates that all individuals have inherent worth, this value does not extend far enough to guarantee its protection. Singer has subsequently argued that there are circumstances (albeit rare) where killing humans, is the right thing to do if it brings about more happiness in the community \cite{singer2011}.

This, of course, is in contrast to the second formula of the categorical imperative that stresses that each individual is intrinsically valuable and is not to be instrumentally subordinated to the well-being, happiness, or pleasure of others. An individual cannot be replaced by another individual even if their existence brings about more happiness. This is because Kantians disagree that the 'good' can be understood as something abstract and impersonal, something that is simply just 'good'. Rather, the good is a relational concept because, as Korsgaard argues, all value is tethered. Happiness or pleasure matters because they are \emph{good-for} individuals~\citep{korsgaard2018fellow}. Therefore, crucially, something can only be considered as universally good when it is \emph{good-for} everyone, from every individual's perspective.
\subsubsection{Critique of Aggregating the Good}
This brings us to the second critique: the issue of aggregation. Because utilitarians do not consider individuals as intrinsically valuable, the moral decision procedure thus allows aggregating the good. There is therefore no prohibition against using individuals, even if they are innocent, as mere means, as long as the loss of their dignity is for maximizing a sufficient amount of ‘good’ in the community. Yet, this aggregation does not make sense, not only because it may violate the second formula of the categorical imperative but also because of the tethered notion of the ‘good’. As Korsgaard contends, “what is good-for me plus what is good-for you is not necessarily good-for \emph{anyone}.”~\cite[p.~157]{korsgaard2018fellow}. Moreover, there is a problem when maximizing the good involves taking something away from an individual: if giving something to Adam brings him more pleasure than it would Eve, the utilitarian would say we ought to give it to Adam. But this action would just be \emph{good-for} Adam and it would be \emph{worse-for} Eve---it is not better for, or \emph{good-for} everyone. The good, because it is tethered, is not something we can aggregate~\citep{korsgaard2018fellow}. This particularly is more concerning when those on the receiving end of aggregation are already marginalized individuals.
\subsubsection{Critique of the Lack of Robustness}\label{subsec:kant.lack.robustness}
The third critique is that the utilitarian approach of calculating outcomes, whether it be maximizing benefit or reducing harm, is not a robust approach to identifying which actions should be promoted or prohibited. As Onora O’Neill points out, circumstances can change the outcome. Take the act of lying as an example: typically, lying is considered to be harmful and immoral and thus, generally prohibited. Yet not all lies, such as ‘white lies’, cause harm. The same goes for telling the truth, which often provides a benefit, but can also inflict emotional harm. Focusing on consequences is unpredictable and, therefore, provides a less robust standard to measure the rightness and wrongness of actions. The link between harmful consequences and actions is unstable. The connection between norms, or duties, and actions is a more robust approach to morally evaluating actions. Duties, such as the duties of civility or to respect the freedom of others, provide a more useful and robust way of evaluating actions \citep{o2022philosopher}. 

Utilitarianism thus has a limitless scope of acceptable actions as it does not depend on the type of action but rather the consequence, or utility, it brings about. Being dishonest for the utilitarian, for example, is not \emph{categorically} condemned; for this reason, it lacks precision as the rightness or wrongness of an action can only be determined until after it is carried out and the consequences are known. The Kantian approach, while it has a more restricted scope given its categorical stance on duties, provides a precise and fundamental standard of action, where moral authority persists irrespective of any potential harmful consequence. Being dishonest (to some Kantians) is always wrong irrespective of unforeseen consequences.

\section{Fairness Metrics for Disparate Treatment and Disparate Impact}\label{sec:fairness}

Transitioning from our review of Kantian deontology, this section shifts focus on to fairness metrics in supervised learning. We integrate fairness metrics within a structured regulatory framework, aiming to facilitate the following discussions. With this transition, we can establish connections with ethical perspectives, laying the groundwork for an examination of disparate treatment/impact and deontology/utilitarianism in algorithmic decision-making systems.

In the field of computer science, researchers have traditionally quantified various statistical disparities that exist among protected attributes. These measures are often labelled using terms from the social sciences~\citep{DBLP:conf/fat/Binns18}, such as equal opportunity~\citep{DBLP:conf/nips/HardtPNS16} and equal treatment~\citep{DBLP:conf/fat/CandelaWHJRAH023}. These notions have subsequently been incorporated into the study of algorithmic fairness to address historical and systemic biases that can arise in decision-making systems.

In order to support our subsequent discussions, we now outline the notation we use to define a subset of metrics which will be central to our position. Let $X$ and $Y$ be random variables taking values in $\mathcal{X}\subseteq\mathbb{R}^d$ and $\mathcal{Y}\subseteq\mathbb{R}$, respectively. A \textit{predictor} is a function $f\colon\mathcal{X}\to\mathcal{Y}$,  also called a model. For conceptual simplicity, the domain of the target feature is $\mathrm{dom}(Y)=\{0, 1\}$ (binary classification). We assume a feature modelling protected social groups denoted by $Z$, called \textit{protected feature}, and assume it to be binary valued in the theoretical analysis.

In alignment with regulatory frameworks like Title VII of the Civil Rights Act of 1964, which prohibit discrimination in employment based on protected characteristics like race and sex~\citep{titleVII1964}, we distinguish between two forms of discrimination: \emph{disparate treatment}, aiming to prevent intentional model discrimination, and \emph{disparate impact}, addressing unintentional effects across subpopulations. The former examines the system's intentions, focusing on model predictions $f_\theta(X)$, while the latter considers effects, interpreted as the error $f_\theta(X)-Y$. This work employs a specific subset of fairness metrics; for a comprehensive overview, refer to \cite{barocas-hardt-narayanan}.

\subsection{Disparate Treatment}\label{subsec:disparate.treatment}

A typical disparate treatment case involves intentional discrimination, necessitating users to demonstrate less favourable treatment motivated by a protected characteristic. To succeed in a disparate treatment claim, users must show an adverse action taken due to their protected characteristics. We interpret fairness metrics based on model predictions $f_\theta(X)$, not model errors $\epsilon = f_\theta(X) - Y$, as group fairness metrics measuring disparate treatment. These include:

\begin{definition}\label{def:dp}\textit{(Demographic Parity (DP))} A model $f$ achieves demographic parity if $f(X) \perp Z$. We can derive an unfairness metric as $d(P(f(X)|Z=z),P(f(X))$, where $d(\cdot)$ is a distance between probability distributions.
\end{definition}


\begin{definition}\label{def:cf}\textit{(Counterfactual Fairness (CF))}.
A model $f_\theta$ is counterfactualy fair if, under any context $X = x$ and $Z = z$,$P(f_\theta^{Z\leftarrow z}(x) = y | X = x, Z = z) = P(f_\theta^{Z\leftarrow z_0}(x) = y | X = x, Z = z)$for all $y$ and for any value $z_0$ attainable by $Z$.\citep{kusner2018}
\end{definition}

Another type of fairness metric that can potentially align with disparate treatment regulatory framework, but does not explicitly rely on the protected attribute, are \textit{individual fairness} metrics, which are grounded in the principle of treating similar individuals equally. These metrics aim to ensure that any two individuals who exhibit similar characteristics in the context of a specific algorithmic prediction task are classified in an identical manner. Metrics such as the Lipschitz condition~\citep{DworkHPRZ2012_IndFair} and similarity graphs between individuals~\citep{DBLP:conf/nips/PetersenMSY21} have been proposed to calculate this in different scenarios.

\begin{definition}\label{def:individual.fairness}\textit{Individual Fairness} A model achieves individual fairness w.r.t. two individual samples if $d (f(x),f(x')) \leq \mathcal{L} \cdot d(x,x')  \forall x, x' \in X$ where $\mathcal{L}>0$ is a constant and $d(\cdot,\cdot)$ is a distance function~\citep{DBLP:conf/nips/PetersenMSY21}
\end{definition}

\subsection{Disparate Impact}\label{subsec:disparate.impact}

In contrast to disparate treatment, disparate impact cases do not necessitate proof of intent as the evaluation shifts towards the discriminatory effects of a seemingly neutral model. Those impacted through the disparate impact account can build a case, for example, by demonstrating that an employment practice substantially affects specific groups. This could manifest in a scenario where a certain job offers more opportunities to black applicants compared to their white counterparts. Employment practices that disparately impact disadvantaged racial groups are deemed unlawful, unless the employer can substantiate their necessity for business reasons~\citep{kim2022raceaware}. We argue that metrics relying on model errors align with disparate impact discrimination, thus a utilitarian approach, and include the following:

\begin{definition}\textit{(Equal Opportunity (EO))}\label{eq:TPR} A model $f$ achieves equal opportunity if $\forall z.\,P(f(X)|Y=1,Z=z) = P(f(X)=1|Y=1)$.~\cite{DBLP:conf/nips/HardtPNS16}\label{def:eo}
\end{definition}

\begin{definition}\textit{(Accuracy Equality (AE))}\label{eq:AE} A model $f$ achieves accuracy equality if $\forall z.\ \forall i \in (0,1)\,P(f(X)=i|Y=i,Z=z) = P((X)=i|Y=i)$.~\cite{wachter2021bias}\label{def:ae}
\end{definition}

An instance where a model violates disparate impact without exhibiting disparate treatment can therefore be illustrated by either a random classifier that uniformly treats all users equally without regard to input data or an algorithm that automatically approves loans for every applicant. In both scenarios, errors may vary across subpopulations, thus resulting in disparate impact even though equal treatment is achieved.

\section{Deontological AI Alignment}

This section considers the compatibility of Kantian deontological ethics with the field of AI fairness and alignment.
We also offer examples of how Kantian ethics can inform AI fairness metrics. At the intersection of these two fields, an existing misalignment has already been noted~\citep{DBLP:conf/aies/FazelpourL20}. In light of this, several research lines have proposed various frameworks to help understand existing notions of algorithmic fairness in moral philosophy terms, e.g. ~\citep{DBLP:conf/fat/HeidariLGK19,DBLP:conf/aies/SimonsBW21}. In this paper, we argue for aligning fairness metrics alongside Kantian deontology.

\subsection{The Issue of Aggregation}\label{subsec:aggregation}
One of the key Kantian critiques charged against utilitarianism is that the utilitarian aggregates the concept of the ‘good’, which within the statistical learning context inevitably involves optimization for the overall outcomes at the expense of one or several individuals' well-being. The intrinsic value of every individual is set aside in the utilitarian context in favour of accumulating a perceived external ‘good’, such as happiness or pleasure (cf. Section~\ref{sec:overview}) that is understood as providing some benefit from the AI applications perspective~\citep{london2023beneficent}.

When considering group fairness, analyzing an individual through group membership, such as treating them based on geographical location or immutable characteristics like ethnicity, and subsequently optimizing or altering decisions based on the group characteristics, erodes the intrinsic value of each individual. If we take individuals to be intrinsically valuable, this means that their value is independent of specific characteristics; we must not see them as mere parts to be subordinated for collective or group ends. We must not, according to Kantian principles, deny an individual's intrinsic self-worth by focusing on specific features to bring about a desired social outcome. The moral law grants impenetrable worth to each individual and, therefore, every individual is to be treated as an autonomous entity equal to everyone else. Giving preference to underprivileged groups may, however, be permitted as the categorical imperative interpreted through the humanity formula rejects indifference to those in vulnerable circumstances to help them maintain the respect and dignity they deserve.

This observation underscores the nuanced challenge of measuring group-level fairness aligning with the preservation of individual rights and dignity, resonating with the Kantian call for how actions affect each individual. AI system evaluation should underscore the importance of designing fairness metrics that truly consider the treatment of all individuals and not just a meeting some decision quota (demographic parity) or achieving similar error rates (equal opportunity) but also look at how each of the attributes being used contribute to the final decision. 


\subsection{Utilitarian AI Fairness Metrics}

This section draws parallels between utilitarian and deontological ethics and the fairness metrics associated with disparate treatment and disparate impact.

Disparate impact metrics are rooted in measuring error disparities among protected groups. Specifically, we align metrics of disparate impact such as \emph{equal opportunity}, Definition~\ref{def:eo} (differences in the True Positive Rates), and \emph{accuracy equality} Definition~\ref{def:ae} (accuracy differences). These metrics hinge on evaluating outcomes, particularly group disparities of model errors. We thus argue that these metrics align with a \emph{utilitarian} approach that focuses on the disparities on the consequences of the model decisions.

A notable technical limitation is that utilitarian metrics (disparate impact fairness) necessitate labelled data for computation. Acquiring labelled data post-deployment poses a considerable challenge and is often impractical,  exhibiting well-known biases, such as confounding, selection, and missingness~\citep{feng2023towards}.\footnote{\emph{Confounding Bias:} Misleading results due to an unconsidered external variable.
\emph{Selection Bias:} Skewed results from non-representative data samples.
\emph{Missingness Bias:} Distortion from systematically absent data~\citep{good2012common}.} Take, for instance, a loan scenario: to compute the True Positive Rate, obtaining False Negatives would be essential. This entails acquiring data for individuals whose model declined a loan but could have repaid it. Privacy concerns, legal constraints, and the dynamic nature of real-world scenarios compound the difficulty in obtaining such post-deployment data~\citep{pearl2016causal,feng2023towards}. In many cases, individuals may not be willing to disclose sensitive information and regulations may limit the collection and use of certain types of data.

In contrast, and in light of our Kantian deontological review, we argue that \emph{deontological principles} prioritize assessing model behaviour rather than an exclusive focus on error rates, therefore, aligning with disparate treatment, metrics such as \emph{demographic parity}, \emph{counterfactual fairness}, and \emph{individual fairness}. These metrics emphasize the \emph{conduct} of AI models and align more closely with the \emph{deontological} approach.

Consider the example of demographic parity Definition~\ref{def:dp}, Kantian ethics would generally contend that an AI model achieves the correct action when its predictions are independent of non-chosen factors at birth (protected attribute), echoing the categorical imperative (humanity formula) to treat individuals as \emph{ends}, never as \emph{mere means} to an end. As mentioned in the introduction, the deontological approach aligns well with widely accepted international declarations that recognize a set of inalienable rights and duties towards the individual, such as the \emph{Universal Declaration of Human Rights}. A similar metric introduced by legal scholars aiming to align with the European Court of Justice \enquote{gold standard} ~\citep{wachter2021fairness} is Conditional Demographic Parity, which extends Demographic Parity by setting one or more attributes fixed (this only differs on the sense that one or more additional covariate conditions are added).

Despite ongoing technical challenges, we argue that the pursuit of ethical AI systems may encounter less technological burden when guided by deontological principles. Unlike utilitarian approaches that rely on post-deployment data, deontological principles provide a framework based on inherent ethical obligations and duties. This perspective is not as contingent on obtaining specific post-deployment data, potentially offering a more straightforward path to ethical alignment in AI systems. With this in mind, if we examine comprehensive surveys of fairness metrics, such as the one by Wachter et al.~\cite{wachter2021bias}, twelve out of the fourteen formalized group fairness metrics rely on model errors (disparate impact), which we argue is utilitarian. There is, therefore, a disproportionate focus on utilitarian fairness metrics. The other two metrics rely on model behaviour, aligning with Kantian deontology and disparate treatment notions as outlined in the previous section~\ref{subsec:aggregation}.


\subsection{Balancing Fairness and Performance Metrics}

The classical utilitarian approach argues that morality is founded upon the search for the \emph{summum bonum}, or the highest form and amount of the good, which, for example, Mill defined as happiness~\citep{mill1966utilitarianism}. In the context of AI fairness, this ethical pursuit can be likened to a secondary optimization endeavour after optimization, akin to fine-tuning, where the goal is to achieve fairness in tandem with optimizing model performance. Techniques such as post-processing adjustments~\citep{DBLP:journals/natmi/RodolfaLG21} or regularization methods~\citep{DBLP:conf/aistats/ZafarVGG17,DBLP:journals/jmlr/ZafarVGG19} are normally used in this secondary optimization stage, aiming to reshape AI systems into less discriminatory entities while maintaining high-performance algorithms. The literature commonly refers to the concept of the \enquote{Pareto-Frontier}, indicating a model status where enhancing accuracy comes at the cost of fairness, and vice versa~\citep{ho2020affirmative}. Any model not situated on the Pareto frontier holds the potential for fairness improvement without sacrificing accuracy, thereby classifying them as potential candidates for improvement~\citep{pmlr-v119-martinez20a}.

This critique finds relevance in contemporary discussions around group-based fairness metrics in AI; when optimizing for fairness there is also a need to ensure the protection of individual rights and dignity. An illustrative example of this concern can be found in Yule's effect on fairness, as highlighted by Ruggieri et al~\cite{DBLP:conf/aaai/Ruggieri0PST23}:

\blockquote{\textit{When pursuing group fairness objectives, such as independence, but failing to account for the control variable Z, fair machine learning algorithms may inadvertently yield disparate effects on distinct distributions. Some subgroups may benefit positively from these fairness efforts, experiencing higher fairness, while others may suffer negative consequences, resulting in lower fairness.}}

However, following the third Kantian critique of utilitarianism (cf Section~\ref{subsec:kant.lack.robustness}), we question the trade-off of maximizing benefit or reducing harm as an approach for determining which actions should be deemed morally permissible or prohibited. This critique reminds us that focusing on the consequences may not provide a sufficiently robust foundation for determining what constitutes as ethical decisions and actions, especially in complex domains like AI fairness. As Kant maintained, individuals ought to be treated as ends in themselves, thus safeguarding their intrinsic worth against any potential outcomes. In the context of AI fairness, we translate this as post-processing techniques that aim to optimize fairness might violate this principled approach. If the AI fairness metric approach wishes to align itself with well-established direction of international standards and rights of the individual, it ought to embrace a Kantian stance within its scope (the one we provide in this paper) and include equality notions that measure what the model has learnt,  focusing on the model decisions rather than the model errors, i.e. the disparate treatment vs disparate impact approach(cf Section~\ref{sec:fairness}).

\subsection{Categorical Imperative as Procedural Fairness}
The Kantian deontological perspective proposes that individuals should act according to the various formulations of the categorical imperative. This approach to morality is thus \emph{procedural}. As Korsgaard argues, the Kantian approach:

\say{
\textit{\ldots thinks that there are answers to moral questions \emph{because} there are correct procedures for arriving at them.}}\cite[p.~36-37]{korsgaard1996sources}

With a deontological approach there is a set of procedural principles or rules, such as the categorical imperative, that provide a systematic method to help us arrive at morally permissible actions.

Applied to AI, this notion encourages developing impartial and equitable ethical procedures across diverse contexts. This aligns with the notion of procedural fairness that has been introduced in fair machine learning literature. Procedural fairness relates to the process, not just the model, which includes the data, processing, and mechanism used to achieve the outcome~\citep{DBLP:conf/fat/MarcinkowskiKSL20,DBLP:conf/aaai/Grgic-HlacaZGW18}.  If we assume that machines are judged against the same standards, their results will be perceived as fair if people understand that the system meets these same criteria.

Focusing on procedure is a familiar approach as it is also integral to many legal systems; not only the outcomes but the integrity of due process often defines the quality of the legal system. For example, while going to prison is an undesirable outcome, if the procedure leading to this decision was rightly followed by the lawyers and judge, the decision is more often than not deemed as fair.

Procedural fairness, as seen through the lens of the categorical imperative, underscores the importance of crafting AI systems that adhere to universal rules and treat all individuals with equal consideration. In practice, this means that the procedures governing AI decision-making should be free from biases, discrimination, and arbitrary distinctions, not just having an unbiased algorithm. By upholding principles that transcend specific circumstances and group identities, AI systems can align with the moral imperative of treating every individual as an end in themselves, rather than as a means to an end. Kant's categorical imperative thus provides a philosophical foundation for procedural fairness in AI ethics, guiding the development of AI systems that operate within a universal framework of equitable and just procedures.

\section{Aligning with Deontological Principles: Use Cases}
In this paper, we have presented an exploration of the intersection between Kantian deontological ethics and the fairness metrics aiming to align with our ethical values. We have argued not only that the current approach to fairness metrics disproportionately favours a utilitarian perspective, to the detriment of not focusing on the conception of the individual that aligns better with individual rights and established international declarations, but that an alignment with a deontological approach is feasible. In this section, we discuss two potential use cases aiming to demonstrate the applicability of Kantian deontological ethics in AI: a loan-granting application using supervised learning and education and technology applications using generative AI.

\subsection{Loan Approval}

In the context of loan approvals, AI systems are typically employed to evaluate the creditworthiness of applicants~\citep{DBLP:journals/ijhci/PurificatoLFL23}. These systems rely on historical data and statistical learning to make decisions and have subsequently raised concerns on the ethical deployment of such technologies~\citep{vecchione2021algorithmic,DBLP:journals/asc/DastileCP20}. A notable example of this issue was brought to light by The Markup in 2021~\citep{martinez2021secret}. Their investigation revealed how a widely used algorithm to help determine mortgage applications in the United States significantly disadvantaged applicants, particularly racial minorities, by exhibiting disparate rejection rates.

The Markup's analysis underscored the role of the algorithm in reinforcing these disparities. Although human factors, such as loan officer bias, play a part, it was evident that the algorithm was a significant contributor to the biased decisions. The study highlighted a specific trend: applicants whose primary income source came from the gig economy faced higher rates of loan refusal.  This trend disproportionately affects racial minorities in the U.S., who are more likely to depend on the gig economy as their primary source of income. The investigation found a significant disparity, with racial minorities being $40-80\%$ more likely to have their mortgage applications denied.  A case in point was a Black couple in North Carolina. Despite having robust financial credentials---high credit scores, adequate income, and savings exceeding the necessary down payment---they were denied a loan on the grounds that one of them was a contract worker~\citep{martinez2021secret}. This decision overlooked their overall financial stability and reduced their evaluation to a single criterion: their employment in the gig economy, which was interpreted as indicative of financial instability.

\paragraph{The Issue of Aggregation}  Traditional loan approval algorithms often optimize for financial metrics, such as relying on credit scores, income levels, and debt-to-income ratios~\citep{DBLP:journals/asc/DastileCP20}. When examined through the lens of Kantian deontology, particularly the \emph{Formula of Humanity}, these practices require closer scrutiny. The first sub-principle of the categorical imperative---avoiding treating others as \emph{\enquote{a mere means to an end}}---is not violated in voluntary bank-customer transactions as banks are only treating consenting applicants as means to their end and can act to ensure that the loans they grant will not default. But the second sub-principle---treating others as \emph{\enquote{ends-in-themselves}}---is morally relevant in this context. Banks can treat applicants as means, but they also must \emph{\enquote{harmonize with}} or attain \emph{\enquote{a positive agreement with humanity as an end in itself}} (4:430-431).

Treating others as ends-in-themselves grants intrinsic worth to each individual.  This approach means looking beyond group affiliations or categories to which an individual may belong. Recognizing the identity traits that make each person unique is crucial, rather than reducing them to mere representatives of these groups. Such a reductionist view not only neglects their unique identity but also undermines their humanity. In the Kantian view, the collective ensemble of these diverse identity traits imbues an individual with impenetrable value. Hence, selecting statistics of individuals by their group characteristics is not only reductive but also ethically flawed, as it fails to honor their complex and singular nature as human beings.

In this loan approval context, \emph{(i)} \textit{Avoiding Group Representations} becomes crucial. Decision-making should not be solely based on group characteristics, including the status as a gig worker or assumptions based on gender or ethnicity. These attributes viewed individually, which are often not up to the individual's choice, are not reflective of an individual’s entire identity and capabilities and could lead to unequal treatment. In the case of the loan through the Kantian deontological lens, each applicant deserves a holistic and equal evaluation, resonating with the principle of respecting each person as an end in themselves. \emph{(ii)} \textit{Beyond Financial Metrics}: An over-reliance on statistical correlations drawn from employment stability, educational background, or community involvement can lead to biases and misrepresentations. These factors, while potentially indicative of financial capabilities, do not capture the entirety of an individual's ability to repay a loan. This is evident in The Markup's case study where the couple, despite having high credit scores, six-figure incomes, and substantial savings, even paying rent higher than their prospective mortgage payments, were narrowly evaluated on their employment status. Such a reductionist approach contravenes Kant's principle of treating individuals as ends in themselves, failing to consider their unique combination of traits that can more collectively determine their creditworthiness.

\paragraph{Utilitarian Metrics and the Challenge of Denied Loan Applications} The ethical complexity of measuring utilitarian fairness metrics in loan scenarios lies in the difficulty of accurately identifying model errors such as false negatives, i.e. individuals who would have successfully repaid a loan but were initially denied by the model. To determine false negatives, it would require extending loans to those previously denied by the AI system, an approach fraught with financial risks, potential biases, and legal implications, in some cases illegal~\citep{feng2023towards}. This paper argues that reliance on metrics that do not depend on post-deployment labelled data may be more effective both in practice and resonate with the principles of Kantian deontology. Such metrics, which focus on the ethical treatment of individuals by the model rather than the consequences of the model predictions, offer a more viable, pragmatic, and deontologically aligned approach. Deontological fairness metrics circumvent the complexities and ethical dilemmas inherent in utilitarian measures that depend on outcomes, particularly in scenarios where gathering accurate post-deployment data is impractical or unethical.
 
\paragraph{Balancing Fairness and Performance Metrics} Financial systems are often designed to maximize financial gain without considering fairness metrics. An alternative perspective~\citep{papageorgiou2023necessity}, in line with German jurisprudence~\citep{BGB,KWG}, which suggests that degrading performance, such as approving loans for initially unqualified applicants, might be legally questionable. Aligning loan approvals with Kantian deontology may not necessarily entail a trade-off between fairness and performance as deontological metrics do not rely on error rates, which do not need changes in the optimization. Recent research by Rodolfa et al. ~\cite{DBLP:journals/natmi/RodolfaLG21} has argued that by modifying the threshold on the model decision, negligible fairness-tradeoff can be obtained,\footnote{It's important to note, especially in contexts like the US and UK, that the use of protected characteristics in decision-making is largely restricted, except under specific, limited conditions~\citep{belz1991equality,fredman2014addressing}} contrary to the premise that ethical principles must always compete with optimization metrics. It might, therefore, be feasible to design AI systems where ethical treatment of individuals and effective financial management are not mutually exclusive goals but complementary objectives. 

\paragraph{Procedural Fairness.} In line with the Kantian emphasis on procedural ethics, the process of how the AI system makes decisions is as, or perhaps even more, important as the decision itself. This involves ensuring that the whole pipeline is up to ethical standards, for example, that the data gathering/collection process meets general data protection requirements, that the training data for the model is non-biased and representative of the population, the algorithm is audited for fairness, and that there are mechanisms in place for applicants to appeal or question decisions~\citep{DBLP:journals/widm/NtoutsiFGINVRTP20}. Therefore, an AI system is designed to adhere to procedural fairness if the whole pipeline meets specific ethical requirements rather than just emphasizing the need to achieve fairness metrics standards (disparate impact). 
\subsection{Students Grading System: Ofqual}

The Ofqual exam grading controversy in 2020 in the UK involved the use of an algorithm to determine final student grades during the pandemic when traditional exam-sitting (A-level exams) was unavailable, and were subsequently criticized for, amongst other reasons, being unfair and biased against students from disadvantaged backgrounds~\citep{DBLP:conf/fat/Benjamin22}. The algorithm was intended to predict student grades based on a combination of historical grade distribution and performance data, student rank in the school from teacher assessments, and previous exam results~\citep{kolkman2020f,heaton2023algorithm}. This, however, led to widespread dissatisfaction among students, parents, and educators due to its apparent systemic biases and inaccuracies. We will discuss this incident through the lens of Kantian deontology.

\paragraph{The Issue of Aggregation} Kantian deontology emphasizes the intrinsic value of each individual, treating them as ends in themselves and not merely as means to an end. In the Ofqual case, the algorithm's reliance on historical data from schools led to individual students being largely judged not on their own merit but based on the performance of their predecessors. Furthermore, the usage of statistical aggregations and features that have a relation with controversial variables, such as the school to which a student belongs, impacted the model. This contradicts the Kantian principle to recognize the intrinsic worth of each individual.

\blockquote{\textit{There is sometimes a danger where you have an exceptionally high-performing child in a low-performing school to be in a situation where they don’t get the grades that they want to.}}~\citep{Adams2020}

This statement by politician Gavin Williamson in an interview with LBC underscores the issue of aggregating individual performances, potentially compromising the well-being of standout individuals simply for belonging to an institution. The algorithm's reliance on past school results failed to recognize each student as a unique individual with distinct potentials and achievements.

\paragraph{Utilitarian Metrics} Public discourse, particularly in social media and news outlets, evaluated the Ofqual algorithm by comparing its predictions with the previous year's results. For instance, The Guardian reported, \enquote{\textit{The proportion of girls in England awarded A or above rose from 25\% to 28\%, overtaking that of boys, who received A or above for 27\% of their entries.}}~\citep{Adams2020}

This evaluation aligns more with a deontological perspective, as it focuses on the fairness of the model's decisions rather than its consequences. A utilitarian approach would require gathering real student data, which was the original problem the machine learning model was trying to solve: avoid examining students in an open classroom due to COVID-19 pandemic restrictions. Additionally, conducting in-person exams might have introduced ethical concerns, as students evaluated algorithmically and those assessed in person would have faced different opportunities, potentially violating principles of equal academic treatment.

\paragraph{Balancing Fairness and Performance Metrics} The Ofqual scenario underscores a critical dilemma in the use of AI techniques for decision-making: the trade-off between algorithmic efficiency and fairness. While the Ofqual algorithm demonstrated efficiency by rapidly processing a vast number of grades, it incorporated historical biases and systemic disadvantages, bringing inherent conflict in the trade-off between performance and fairness. In line with Kantian ethics, the trade-off between performance and fairness cannot be the sole criterion for ethical validation in AI applications, raising a fundamental question: \emph{should AI techniques even be employed in student grading contexts?} This is especially pertinent given that after public outrage, students instead received grades solely based on their teacher's estimate of what their grade would have been. The Ofqual experience serves as a cautionary tale, highlighting the need for a reflection on whether the usage of AI systems for automation actually improves societal aspects, even when considering \enquote{fairness metrics}.

\paragraph{Categorical Imperative as Procedural Fairness} Kantian deontology emphasizes the value of process over outcomes. Procedural fairness, within AI, extends beyond the statistical difference in the model predictions (fairness metrics), but requires that the full process of making predictions respects a set of ethical principles. For example, fairness metrics that focus solely on the consistent functioning of an algorithm with respect to two protected groups may overlook crucial aspects of discrimination or bias in the step data gathering or data processing. A principle like the categorical imperative compels us to consider that a system is ethically engineered if each of the steps needed towards making a prediction \citep{mle,DesignMLSystems} are grounded in moral principles. Within AI research, while questions of distributive justice, such as equitable outcomes, receive significant attention, procedural justice is often relegated to a secondary status. From a Kantian perspective, however, ensuring procedural justice is paramount. This involves scrutinizing the data used for training algorithms, the assumptions and values embedded in these systems, and the transparency and accountability of the algorithmic processes. It is about ensuring that these systems are not only fair in what they distribute (outcomes), but also how they operate (processes).

\section{Conclusion}
This paper examined the compatibility of Kantian deontology with AI alignment with fairness metrics. We first revisited the Kantian critique of utilitarianism, which prioritizes outcomes, and explored its relevance to AI fairness. The Kantian framework emphasizes treating individuals as ends in themselves, aligning with the foundational principles of fairness in AI that seek equitable and just treatment for all, irrespective of group identities and model errors.

The current landscape of AI fairness metrics, however, primarily focuses on optimizing outcomes and reducing disparities in model errors, which is more in line with utilitarian principles. The longstanding Kantian critique of this approach reminds us of the need for a deeper examination of the moral principles underlying fairness in AI. The deontological approach also supports a procedural fairness framework for AI ethics, highlighting the importance of impartial and equitable procedures in AI decision-making. This aligns with the established concept of procedural fairness in AI ethics, ensuring that AI systems uphold universal principles of equitable treatment and respect for individual dignity. By integrating Kantian ethics into AI alignment, we not only bring in a widely-accepted prominent moral theory but also strive for a more morally grounded AI landscape that better balances outcomes and procedures in pursuit of fairness and justice.


\paragraph{Researcher Positionality Statement} Our backgrounds and experiences coming from Western education significantly influenced the trajectory of this work. As interdisciplinary researchers, our diverse perspectives enriched the examination of Kantian deontological ethics in the context of AI fairness. Drawing from interdisciplinary backgrounds, including philosophy and computer science, we integrated moral theories into technical domains. Acknowledging our own positions, we recognize the importance of context-specific interpretations of Kantian ethics in different sociocultural settings.

\paragraph{Adverse Impact Statement} While we believe our exploration of Kantian deontological ethics in AI alignment fairness metrics contributes valuable insights, we acknowledge the potential adverse impacts of our work. One unintended consequence may be the oversimplification of ethical considerations in AI systems, as Kantian ethics may not comprehensively address the dynamic and evolving nature of technological landscapes. This oversimplification could lead to a rigid application of ethical principles, potentially hindering the adaptability of AI systems in diverse contexts. Moreover, the dissemination of our findings may inadvertently contribute to a narrow focus on deontological approaches, therefore overshadowing the importance of a holistic ethical framework that integrates multiple perspectives. We caution against the uncritical adoption of our proposed framework and urge for ongoing dialogue and adaptation to ensure that AI ethics remains responsive to emerging challenges and ethical considerations.

\bibliography{references}
\bibliographystyle{apalike}

\end{document}